\documentclass{article}

\usepackage{arxiv}

\usepackage[utf8]{inputenc} 
\usepackage[T1]{fontenc}    
\usepackage{hyperref}       
\usepackage{url}            
\usepackage{booktabs}       
\usepackage{amsfonts}       
\usepackage{nicefrac}       
\usepackage{microtype}      
\usepackage{lipsum}		
\usepackage{graphicx}
\usepackage{doi}
\graphicspath{{Figures/}}
\usepackage{amsmath,amssymb,amsfonts}
\usepackage{algorithmic}
\usepackage{graphicx}
\usepackage{placeins}

\title{The effect of variable labels on deep learning models trained to predict breast density}

\author{Steven Squires \\
	University of Manchester\\
	Manchester, UK
	\And	
	Elaine F. Harkness \\
	University of Manchester\\
	Manchester, UK
	\And	
	D. Gareth Evans \\
	University of Manchester\\
	Manchester, UK
	\And
	Susan M. Astley \\
	University of Manchester\\
	Manchester, UK
}

\renewcommand{\shorttitle}{\textit{arXiv} Template}

\hypersetup{
pdftitle={A template for the arxiv style},
pdfsubject={q-bio.NC, q-bio.QM},
pdfauthor={David S.~Hippocampus, Elias D.~Striatum},
pdfkeywords={First keyword, Second keyword, More},
}

\begin{document}
\maketitle

\begin{abstract}
\noindent {\bf Purpose: High breast density is associated with reduced efficacy of mammographic screening and increased risk of developing breast cancer. Accurate and reliable automated density estimates can be used for direct risk prediction and passing density related information to further predictive models. Expert reader assessments of density show a strong relationship to cancer risk but also inter-reader variation. The effect of label variability on model performance is important when considering how to utilise automated methods for both research and clinical purposes.}   \\
{\bf Methods: We utilise subsets of images with density labels from the same 13 readers and 12 reader pairs, and train a deep transfer learning model which is used to assess how label variability affects the mapping from representation to prediction. We then create two end-to-end models: one that is trained on averaged labels across the reader pairs and the second that is trained using individual reader scores, with a novel alteration to the objective function. The combination of these two end-to-end models allows us to investigate the effect of label variability on the model representation formed.}  \\
{\bf Results: We show that the trained mappings from representations to labels are altered considerably by the variability of reader scores. Training on labels with distribution variation removed causes the Spearman rank correlation coefficients to rise from $0.751\pm0.002$ to either $0.815\pm0.006$ when averaging across readers or $0.844\pm0.002$ when averaging across images. However, when we train different models to investigate the representation effect we see little difference, with Spearman rank correlation coefficients of $0.846\pm0.006$ and $0.850\pm0.006$ showing no statistically significant difference in the quality of the model representation with regard to density prediction.}  \\
{\bf Conclusions: We show that the mapping between representation and mammographic density prediction is significantly affected by label variability. However, the effect of the label variability on the model representation is limited. }  \\
\end{abstract}

%
%
%
%
%

\section{Introduction}

Mammographic density is an important factor when assessing women for their risk of developing breast cancer~\cite{boyd2010breast,huo2014mammographic}. There are several different methods of assessing density; one that shows a strong relationship with cancer risk is Visual Analogue Scale assessment (VAS) by radiologists and advanced practitioner radiographers~\cite{astley2018comparison}, specifically the average score assigned by two independent readers, however the process of obtaining these is time-consuming and not scalable. Automated models that could produce similar scores would enable broader use of this measure as a risk prediction tool.

A number of automated methods have been developed to assess breast density~\cite{fonseca2015automatic,kallenberg2016unsupervised,highnam2010robust,keller2012estimation}. Deep learning approaches, both from scratch~\cite{ionescu2019prediction}, and using transfer learning~\cite{squires2020automatic,squires2022automatic}, have been shown to produce density estimates that correlate well with VAS scores.

The reason for the high performance of VAS scores for assessing risk compared with automated volumetric density measures is hypothesised to be that the domain experts are using their professional judgement rather than a more mechanical approach which determines the quantity of dense tissue. The experts are perhaps taking into account the complex interplay of areas of dense and non-dense tissue, including the pattern and distribution, which other methods may not consider. Or perhaps they are identifying other signs such as clusters of classifications~\cite{alsheh2019association}. The logic of training on the assessments of these experts is to utilise their years of experience to train models to make similar predictions. Deep learning is particularly appropriate for VAS prediction because deep models automatically extracts features that might not be apparent to the conscious mind of humans. However, a significant issue with deep learning models is that they are generally considered to be ``black boxes"~\cite{castelvecchi2016can} where we do not know why they are making their predictions. Consequently we either need to be confident in the quality of the model predictions or provide explainable outputs~\cite{amann2020explainability}.

An issue with the assessment of the quality of model prediction is that expert estimates show both intra- and inter-reader variability~\cite{sergeant2013same,sprague2016variation,sperrin2013correcting}. Training on variable data may result in models that produce inaccurate or unreliable predictions. In addition, it was shown that with variation in reader scores it becomes difficult to assess the quality of predictive models because a ``optimal" model would also show a relatively large error~\cite{squires2022automatic}. This difficulty makes it hard to decide which models to choose or to have confidence in the quality of the model performance.

If model predictions are inaccurate women may be assigned to an incorrect risk category, potentially resulting in missed opportunities to perform earlier interventions or to undertake additional investigations, or causing undue alarm. In addition, density scores can be used for other purposes such as estimating the risk of lesions being concealed by dense tissue (masking)~\cite{mainprize2016quantifying,mainprize2019prediction} or as inputs to other automated models, in which case inaccurate or unreliable models would weaken their quality. It is therefore important to investigate the effect of variability in reader labels on model predictions.

The variability of radiologist density scores has been investigated previously which showed that variation between readers could be substantial~\cite{sergeant2013same,sprague2016variation}. There has also been work done to correct scores between different readers~\cite{sperrin2013correcting}. However, these pieces of work were interested in the variability of the scores themselves, not the effect that this variability has on deep learning-based models. 

Deep learning models can be considered as a combination of a model which produces a representation in some feature space together with a mapping module which converts that feature space into a final prediction. Variability in labels may affect either, or both, of these aspects of deep models, and can make a significant difference to how models should be assessed and deployed.

In this work we first adopt a feature extraction and linear regression approach to explore the importance of label variability on the mapping from a feature space to a density score. This methodology allows us to train and test on subsets of the data which enables us to reduce the variation in the labels and assess how the performance of the model changes.

We also investigate the effect of training using variable labels on the quality of the representation created by the deep models; this requires training models end-to-end rather than using them as feature extractors. We create two fully trained models. One is trained on the averaged labels across all the readers and will include the variability due to differences in reader assessment. The second is trained with predictions made separately on the individual reader scores, by a novel alteration to the objective function. This model should be less affected by reader variability in comparison with the model trained on averaged scores.

\section{Materials and Methods}
\label{sec:methods}

\subsection{Reader variability}
\label{sec:reasonForScoreVariation}

Our overall aim is to investigate how variability in reader density labels effects the representation and mapping of our models. Variability can affect the quality of the model through the impact on training and the assessment of model performance. We hypothesise that reader variability has three aspects:

\begin{enumerate}
\item \textbf{Fundamental differences in assessment of images:} readers have a genuinely different assessment of the density of an image. We know that individual reader assessments differ from one another~\cite{sergeant2013same} and also that the average of two reader assessments of density differ from the outputs of automated methods~\cite{astley2018comparison}. Understanding these differences may enable us to design more accurate and reliable models for density-based risk assessment by facilitating understanding of image features associated with increased risk.
\item \textbf{Different distributions:} readers may record different absolute density values but generally (within some random error margin) agree on the order of the images from low to high density. We might expect two readers which had been exposed to different density distributions to produce different scores but still to monotonically increase together (accepting the random errors). For example, if one reader sees very few high density mammograms they might assign a higher density to images than a reader who sees a lot of high density images, due to a tendency to fill up the density space.
\item \textbf{Random errors/uncertainty:} we define random error as the difference left over between two image density assessments after accounting for the distribution and fundamental differences in density prediction. These errors should not be systematic. 
\end{enumerate}

\subsection{Data-set and pre-processing}
\label{sec:explainData}

The dataset comprises sets of four mammographic images for each woman corresponding to the four standard radiographic projections used in screening: RCC, LCC, RMLO, LMLO. Images were all produced by GE Senographe Essential mammogram systems. Two independent expert readers assigned on a pragmatic basis from a pool of 19 readers assessed each image as part of the Predicting Risk Of Cancer At Screening (PROCAS) study~\cite{evans2012assessing} and marked an assessment of density for each radiographic projection on a 10cm Visual Analogue Scale. We refer to these as VAS density scores and they are in the range 0-100\%. 

For our first method, using feature extraction with linear regression, we investigate how well models perform when trained on scores from individual readers and paired reader subsets of the data. To improve reliability, data from individual readers who scored the density for fewer than 4,000 images (1,000 women) were excluded, leaving data from a total of 13 readers. Data from pairs of readers who jointly scored density for fewer than 4,000 images were also excluded. The number of images read by each of these readers and reader pairs are detailed in Tables~\ref{tab:resultsTab4} and~\ref{tab:resultsTab5}, respectively.

The second method is designed to investigate the effect of label variability on the model representation in feature space. This approach uses end-to-end training; we split the whole data-set into training (80\%), validation (10\%) and testing (10\%) sets by woman so that no images from the same woman are in more than one partition. 

In addition, a subset of the complete data-set contains a case control set with images from 338 women. Each of these women are matched with three women without cancer. By controlling for known risk factors we can then produce an estimate regarding the effect of the mammographic density on cancer risk~\cite{astley2018comparison}. None of these women were included in the training and validation partitions. We report the quality of the cancer risk predictions made by all the models produced here.

For pre-processing we scale the images down to a size of $224\times 224$, flip left sided images to the right, limit the pixel value range, invert the pixels, perform histogram equalization and set the pixel range between 0 and 1. This approach has been described previously~\cite{squires2022automatic}.

\subsection{Method 1: transfer learning subset data models for the investigation of model mapping}

For this method we consider the effect of training on the subsets of the data from either an individual reader or the same pair of readers. For clarity, by individual readers we mean that the images included in training are all read by the same reader and the labels used are the labels provided by that reader. By paired readers we mean the set of images read by both readers and the label is the average of the two reader scores. Due to the relatively small amount of data we need models that will produce accurate results but require only a modest number of images. In previous work it was shown that using a feature extractor with a linear regression produces reasonable density predictions~\cite{squires2022automatic}. While a multi-layer perceptron trained on top of the deep feature extractor produced improved results over linear regression differences were modest. An additional advantage of using a linear regression model trained on top of extracted features is that the results are entirely deterministic - they will be identical each time and are optimal for a linear combination of those features (under the mean squared error metric and with that level of regularisation). As we are interested in investigating the effect of variability of the reader estimates it is important to have a consistent model so that we can be confident in the conclusions we draw.

\begin{figure}[ht] 
\begin{center}
\includegraphics[height=3.1cm]{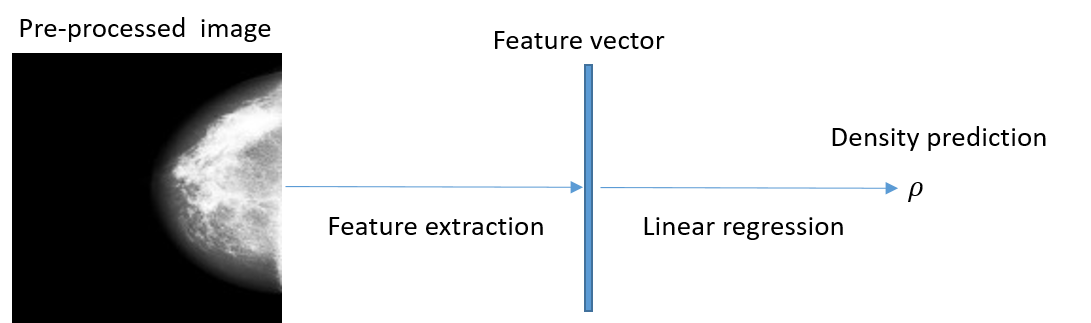}
\end{center}
\caption 
{ \label{fig:diagram1}
A schematic diagram showing the steps in the prediction procedure in the inference stage. Feature extraction involves the application of the pre-trained DenseNet model, with the fully connected part of the network removed, to a pre-processed mammogram image producing a 2,208 feature vector for that image. We add a bias term to make it a 2,209 dimensional vector. The linear regression step applies the learnt weights to the feature vector to output a final density prediction, $\rho$.}
\end{figure}

Drawing on previous work~\cite{squires2022automatic} we utilise the DenseNet~\cite{huang2017densely} model trained on ImageNet~\cite{deng2009imagenet} to produce feature vectors and then train a linear regression to produce density predictions. Pre-processed images (see Section~\ref{sec:explainData}) are fed through a truncated DenseNet model where the final fully connected layer has been removed. The DenseNet model then outputs a 2,208 dimensional feature vector for each input image. We add a bias term (making a 2,209 feature vector) and then use linear regression to produce a density estimate using reader labels as the target. In Figure~\ref{fig:diagram1} we show a schematic version demonstrating the approach for the inference stage.

To make our density predictions we utilise the pseudo-inverse to calculate weights:
\begin{equation}
\textbf{w}=(\textbf{X}_{tr}^T\textbf{X}_{tr}+\lambda_2\textbf{I})^{-1}\textbf{X}_{tr}^T\textbf{y}_{tr}
\end{equation}

\noindent where $\textbf{X}_{tr}\in \mathbb{R}^{n\times p+1}$ is a matrix formed by stacking the $p$-dimensional feature vectors (with the addition of the bias term) produced by the application of the DenseNet feature vector to the training images, $\lambda_2$ is a parameter controlling the $L_2$ regularisation, $\textbf{I}$ is the identity matrix and $\textbf{y}_{tr}\in \mathbb{R}^{n}$ is a vector containing the reader labels. To perform inference on the testing data we apply $\boldsymbol{\hat{y}}_{ts}=\textbf{X}_{ts}\textbf{w}^T$ where $\textbf{X}_{ts}$ is a matrix of the stacked feature vectors from the testing set.

As we are using deep transfer learning with a model trained from ImageNet there is no need to retrain the feature extractor model. The only part of the model we need to train at each point is the linear regression. In terms of the computation we have a matrix $\textbf{X}$ containing the $n$ DenseNet derived feature vectors for all our images. Alongside that we have a $\textbf{Y}\in\mathbb{R}^{n\times2}$ matrix which contains all the pairs of reader scores for all the images. Our approach requires us to sample the correct training and testing data from the matrices and find the appropriate $\textbf{w}$. On a standard desktop computer the training process takes seconds for the entire data-set and less for any of the subsets.

Throughout most of this work we train our models using five-fold cross-validation to make predictions. Thus our predictions are actually made using five different models with the predictions reported as those on the held-out sets. This enables us to make predictions on all sub-sets of the data and although it does add a small amount of additional variation, due to the slightly different training sets, this is not enough to alter our conclusions. 

We produce predictions on the whole data-set, by training using both the averaged reader scores as well as individual reader scores. These provide a baseline to assess the performance of models trained on the subsets of the data.

We can then examine the effect of training and testing on the individual reader scores and pairs of readers. For the former we take just those images with scores by one individual reader and produce a model which is trained using image/label pairs from that reader alone. For pairs of readers, we train models on the images and averaged scores for the pair. We present results after combining the predictions on the CC and MLO images. The differences in prediction between CC and MLO are generally small.

When we train, and test, on individual reader scores we are reducing both the variation due to the fundamental density assessment and also the distribution (points 1 and 2 in Section~\ref{sec:reasonForScoreVariation}). We consider this to be a reduction rather than a removal as individual readers may alter their judgement over time. When utilising paired reader scores we are reducing variations due to all three reasons discussed in Section~\ref{sec:reasonForScoreVariation} although the extent of the reduction is not certain. Using different subsets of the data enables us to investigate the mapping effect and to seek insight into how much of the error is real - due to the model making predictions that differ from the true value.

\subsection{Method 2: end-to-end deep models to investigate the representation formed}
\label{sec:method2}

The possible effect of variable reader scores on a representation formed by a deep model can be illustrated with an example. If we take two images which have similar true density but they have been labelled by different pairs of readers then they may have significantly different scores assigned to them. When we train on these two images the model may fluctuate in training as the model parameters are pulled in different directions trying to make the model learn the different labels. Alternatively, the model may be forced to learn noise to make a good prediction on the training set. This variability could potentially have a significant effect on the quality of the representation formed.

To create a model which should be fully affected by label variability we utilise a ResNet-18~\cite{he2016deep} implementation pretrained on ImageNet~\cite{deng2009imagenet}. The final fully connected layer that maps to 1,000 output neurons is replaced with a layer that maps to one output neuron. The model is trained using the Adam~\cite{kingma2014adam} optimiser with different learning rates. We augment our images by: flipping left-right with a 50\% probability, random rotations and the addition of Gaussian noise. As the ResNet architecture requires three input channels we copy the greyscale image (after augmentation so all three images are identical) across the three channels. The training, validation and testing split was discussed in Section~\ref{sec:explainData}. We train on the combination of the CC and MLO images, selecting the final model by the root mean squared error performance on the validation set. All results are reported on the testing set which consists of 15,291 images. We call this model the \textit{single-predictor}.

\begin{figure}[ht] 
\begin{center}
\includegraphics[height=3.1cm]{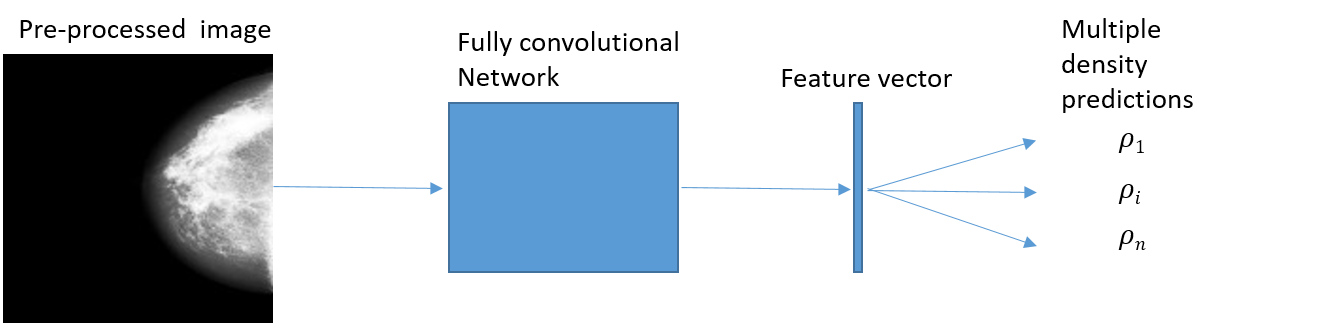}
\end{center}
\caption 
{ \label{fig:diagram2}
The architecture of the multi-predictor model to enable training on all the individual reader scores. 
}
\end{figure}

Our second end-to-end model, the \textit{multi-predictor}, is aimed at reducing the effect of the label variability on the representation formed. A schematic is shown in Figure~\ref{fig:diagram2}. The fully convolutional network is taken from the ResNet~\cite{he2016deep} model with the weights pretrained on ImageNet~\cite{deng2009imagenet} and the final fully connected layers removed. The output of the convolution part of our model is flattened and concatenated to produce a 512-dimensional feature vector; up to this point this model is identical to the previous \textit{single-predictor} model. This feature vector is then linearly mapped to $m=13$ output neurons each which represent one reader score.  The model is trained end-to-end with the linear mapping included in the adjusted model. By training with $m$ output neurons, we can let the model adjust for the label variability by adjustments in the linear weights of the last layer of the model.

For an individual image we have two labels to compare with our two predicted outputs but 11 unknown labels. The entire convolutional part of the model (see Figure~\ref{fig:diagram2}) is updated for each image; but in the final layer only the linear weights of the neurons with known labels are updated. Some of the variation between the different readers will be accommodated by that linear part without affecting the rest of the model - the model should not be pulled different ways by the variability of the labels. However, we still get a signal for all of the images which can be used to adjust the fully connected part of the network. 

To achieve this we utilise mean squared error as a loss function with an additional filter which prevents errors from output neurons with unknown labels from backpropagating through the network. Our prediction output is a $m$-dimensional vector, $\boldsymbol\rho\in\mathbb{R}^m$, where $m$ is the total number of readers across the entire data-set (13 in our case). We then create a $m$-dimensional vector that is a binary filter, $\boldsymbol\phi\in\mathbb{R}^m$. The vector of labels, $\textbf{d}\in\mathbb{R}^m$ is given by a $m$-dimensional vector where each element is a density score from a specific reader - all but two of which are unknown. Our model will make a prediction for all $m$ output neurons and this filter will remove those from the objective function for which there is no label. For an individual image then we have a loss given by:

\begin{equation}
\text{loss}= \sum_i^m[\phi_i(\rho_i-d_i)^2]
\end{equation}
\noindent where $\phi_i$ is 1 if a label is known and 0 otherwise. The filter therefore prevents an error contribution from those outputs that do not have labels.

For each image, only the weights in the final layer associated with the known labels will be updated. However, as the convolutional layers all feed into the fully connected layer the entirety of the fully convolutional layers will be updated for all the images. Therefore, we update the fully convolutional layers with the entire training set whilst updating the linear layers with only the labelled data.

The training procedure we follow is identical to that of the \textit{single-predictor}. This should ensure the only difference in outcomes is due to the methodological differences, not because of differences in the training process or the base-model.

The purpose of these two end-to-end models is to investigate the effect of the label variability on the quality of the representation formed. To do so we need to map the representations to the reader scores. We utilise the same approach as for Method 1 to map the representations to the final scores. We remove the final fully connected layer resulting in a 512 dimensional feature vector. Linear regression is then applied in the same manner as method 1.

\subsection{Comparisons and metrics}
\label{sec:comparisons}

We utilise two metrics to compare to the labels: Root Mean Squared Error (\textit{RMSE}) and Spearman rank correlation (\textit{Rank Corr.}), the former to give absolute differences and the latter to give rank differences. We also use a case-control set to evaluate odds-ratios of developing cancer~\cite{astley2018comparison}.

We can compare to either averaged pairs of reader scores or individual scores. Most of our uncertainties are found via bootstrapping and are reported at the 95\% level. Where we have used other approaches they are discussed as the results are reported. 

\section{Results}

\subsection{Method 1: transfer learning subset data models for the investigation of model mapping}

\begin{table}[ht] 
\centering
\begin{tabular}{|ll|cc|}
\hline
\textbf{Tr} & \textbf{Ts} &  \textbf{Rank Corr.} &   \textbf{RMSE}  \\
\hline
\textbf{Av} & \textbf{Av} &  $0.820\pm0.002$ &   $9.38\pm0.04$ \\
\textbf{Ind} & \textbf{Ind} &  $0.751\pm0.002$ &  $11.63\pm0.05$  \\
\textbf{Av} & \textbf{Ind} &  $0.752\pm0.002$ &  $11.62\pm0.04$  \\
\textbf{Ind} & \textbf{Av} &  $0.818\pm0.002$ &   $9.44\pm0.04$  \\
\hline
\end{tabular}
\caption{Global metrics training (\textbf{\textit{Tr}}) and testing (\textbf{\textit{Ts}}) on averaged (\textbf{\textit{Av}}) and individual (\textbf{\textit{Ind}}) labels. Uncertainty is reported at the $95\%$ level and found via bootstrapping with 1,000 repeats.}
\label{tab:resultsTab2}
\end{table}

The prediction metrics on the whole dataset (via cross-validation), of 76,973 CC images and 77,080 MLO images are shown in Table~\ref{tab:resultsTab2}. The metrics are Spearman's rank correlation coefficient (\textbf{\textit{Rank Corr.}}) and root mean squared error (\textbf{\textit{RMSE}}). The uncertainty is at the $95\%$ level produced by performing 1,000 sets of bootstrapping. We show the results found by training (\textit{\textbf{Tr}}) on the averaged labels (\textit{\textbf{Av}}) and the individual labels (\textit{\textbf{Ind}}) along with testing on the averaged labels and the individual labels (\textbf{\textit{Ts}}), which gives us four sets of prediction metrics in total. There are large differences in model performance between the individual and averaged test sets. However, there is a much smaller variation due to using individual or averaged labels for the training set. We note that a similar finding was made previously~\cite{squires2022automatic} including when they used a non-linear MLP after the feature extraction, so this is not due to the linear regression. The main purpose of these results though is to provide global metrics for comparison when we investigate the effect of training on the individual and paired values.

\begin{table}[ht] 
\centering
\begin{tabular}{|c|cccc|}
\hline
\textbf{Image} & \multicolumn{4}{|c|}{\textbf{Train-Test}} \\
\textbf{number} & \textbf{Av-Av}& \textbf{Ind-Ind}& \textbf{Av-Ind}& \textbf{Ind-Av} \\
\hline
20,000 & $9.54\pm0.08$ & $11.80\pm0.11$ & $11.73\pm0.08$ & $9.67\pm0.08$ \\
10,000 & $9.73\pm0.11$ & $12.03\pm0.16$ & $11.93\pm0.11$ & $9.92\pm0.11$ \\
7,000  & $9.79\pm0.13$ & $12.07\pm0.19$ & $11.93\pm0.13$ & $10.02\pm0.14$ \\
4,000  & $10.05\pm0.18$& $12.22\pm0.24$ & $12.08\pm0.16$ & $10.28\pm0.17$ \\
2,000  & $10.33\pm0.26$& $12.79\pm0.36$ & $12.51\pm0.23$ & $10.67\pm0.27$ \\
\hline
\end{tabular}
\caption{RMSE for different sizes of the data-set. The number of images refers to the number of CC or MLO images therefore the number that go into the model before cross-validation takes place.}
\label{tab:resultsTab3}
\end{table}

When we train our models on the paired and individual reader subsets we reduce the amount of data which we would expect to result in poorer predictive performance. We therefore produce a set of metrics which gives us an estimate of the predictive quality when the data-set is smaller. In Table~\ref{tab:resultsTab3} we show the RMSE for different sizes of the data-set, we see a steady reduction in predictive quality with smaller numbers of images. The numbers refer to the numbers of CC or MLO images.

\begin{table}[ht] 
\centering
\begin{tabular}{|l|c|cc|}
\hline
\textbf{Reader} & \textbf{Number} & \textbf{Rank Corr.} &  \textbf{RMSE}  \\
\hline
A  &    2263 &  $0.808\pm0.012$ &	$11.01\pm0.30$ \\
B  &   10821 & $0.817\pm0.006$ &	$8.49\pm0.17$ \\
C  &   19262 &   $0.815\pm0.003$&	$8.57\pm0.09$ \\
D &   27829 &   $0.806\pm0.004$&	$11.32\pm0.09$ \\
E  &   20022 &    $0.796\pm0.004$	&$8.89\pm0.09$ \\
F  &   26239 &    $0.798\pm0.004$&	$10.29\pm0.12$ \\
G &   11469 &   $0.866\pm0.004$	& $9.43\pm0.11$  \\
H  &    2606 &  $0.816\pm0.009$&	$12.83\pm0.28$ \\
I  &    4467 &  $0.832\pm0.008$	&$8.13\pm0.20$ \\
J &    8915 &  $0.771\pm0.01$&	$10.54\pm0.24$  \\
K &    3249 &  $0.856\pm0.007$&	$6.11\pm0.18$  \\
L  &   14137 &   $0.834\pm0.004$&	$10.88\pm0.11$ \\
M  &    4230 &  $0.786\pm0.008$&	$8.37\pm0.15$ \\
\hline
Av1 &  155,509 & $0.815\pm0.006$	&$9.60\pm0.16$ \\
Av2 &  155,509 &   $0.844\pm0.002$	&$9.90\pm0.04$  \\
\hline
All & 77,027 &   $0.751\pm0.002$ &  $11.63\pm0.05$ \\
\hline
\end{tabular}
\caption{Metrics when trained and assessed on individual reader scores. We also show the approximate number of images the models are trained on. \textit{Av1} is the average across the reader metrics and \textit{Av2} is the average across all the images. The averages are substantially better than the overall metrics we find when we trained on the entire data-set (Tables~\ref{tab:resultsTab2} and~\ref{tab:resultsTab3}). \textit{All} shows the global metrics trained on the entire data-set (\textbf{Ind}-\textbf{Ind} from Table~\ref{tab:resultsTab2})}
\label{tab:resultsTab4}
\end{table}

Tables~\ref{tab:resultsTab2} and~\ref{tab:resultsTab3} show global metrics for comparative purposes. In Table~\ref{tab:resultsTab4} we show the quality of the predictions when trained and assessed on only those images labelled by the 13 individual readers (denoted as A-M) and with that individual's density score as the label for each image. We also show the approximate (as the number of CC and MLO are not exactly equal) number of images (\textbf{\textit{Number}}) that each reader has seen. The table also shows the average across the readers as $Av1=\frac{1}{n_r}(\sum_{i=1}^{n_r} \mu_i)$, where $n_r$ is the number of readers and $\mu_i$ the metric of the $i^{th}$ reader. The given uncertainty for $Av1$ is the average of the uncertainties of the individual readers. The metric $Av2$ is an average across all the images which includes repeated image-score pairs. The uncertainty for $Av2$ was found via bootstrapping and is not directly comparable to the uncertainty of $Av1$. The comparison we are interested in is with the quality of the global predictions on the individual (\textbf{Ind}-\textbf{Ind}) shown in Table~\ref{tab:resultsTab2} (also shown as \textit{All} in this table for ease of comparison). The predictions show most models getting significant improvements in performance. If we factor in the smaller size of the training set, by considering Table~\ref{tab:resultsTab3}, the gap in effective performance is even higher.

\begin{table}[ht] 
\centering
\begin{tabular}{|l|l|ll|}
\hline
{} & \textbf{Number} & \textbf{Rank Corr.} & \textbf{RMSE} \\
\hline
\textbf{B\_E}   &   3250 &  $0.818\pm0.009$	&$8.18\pm0.22$ \\
\textbf{B\_F}   &   2655 &  $0.843\pm0.012$	&$8.23\pm0.29$ \\
\textbf{C\_D}  &   9764 &   $0.86\pm0.004$	&$8.11\pm0.10$ \\
\textbf{D\_E}  &   2581 &  $0.848\pm0.008$	&$8.60\pm0.190$ \\
\textbf{D\_F}  &   4087 &  $0.85\pm0.007$	&$8.98\pm0.18$ \\
\textbf{D\_J} &   2379 &  $0.819\pm0.01$	&$9.82\pm0.26$  \\
\textbf{D\_L}   &   2245 &  $0.863\pm0.008$	&$9.33\pm0.22$ \\
\textbf{E\_F}   &   4088 &  $0.841\pm0.007$	&$7.25\pm0.15$ \\
\textbf{E\_G}  &   2523 &  $0.864\pm0.007$	&$8.19\pm0.20$  \\
\textbf{E\_L}   &   2202 &  $0.845\pm0.009$	&$8.37\pm0.20$ \\
\textbf{F\_G}  &   2719 &  $0.861\pm0.007$	&$8.35\pm0.22$  \\
\textbf{F\_L}   &   5254 & $0.864\pm0.005$	&$8.56\pm0.15$ \\
\hline
\textbf{Av1}     &  43747 & $0.848\pm0.008$	& $8.50\pm0.20$ \\
\textbf{Av2}     &  43747 &   $0.864\pm0.002$ &	$8.42\pm0.06$ \\
\hline
\textbf{All}     &  43749 &  $0.834\pm0.002$ & $9.19\pm0.05$ \\
\hline
\end{tabular}
\caption{Metrics for pairs of readers when trained and assessed on the averaged scores of only those pairs. \textit{All} shows the metrics when training and assessing on only those images included in the paired data set.}
\label{tab:resultsTab5}
\end{table}

The results in Tables~\ref{tab:resultsTab2} and~\ref{tab:resultsTab3} demonstrate that assessing and, to a lesser extent, training on the averaged labels produces better model performance than using individual labels. We therefore also use the paired set of data for training and assessing and show metrics resulting from this process in Table~\ref{tab:resultsTab5}. The metrics associated with \textit{Av1} and \textit{Av2} are the same as in Table~\ref{tab:resultsTab4}. We also include an additional set of metrics called \textit{All} which is the metrics when we train together on all the included paired data, i.e. it is another global metric of all the image and scores together but excluding those with fewer than 4,000 images read between them. We see the same pattern as when we trained and tested on the individual reader scores: that we find significantly improved performance compared to training on all the data together.

\begin{figure}[ht] 
\begin{center}
\includegraphics[height=9.8cm]{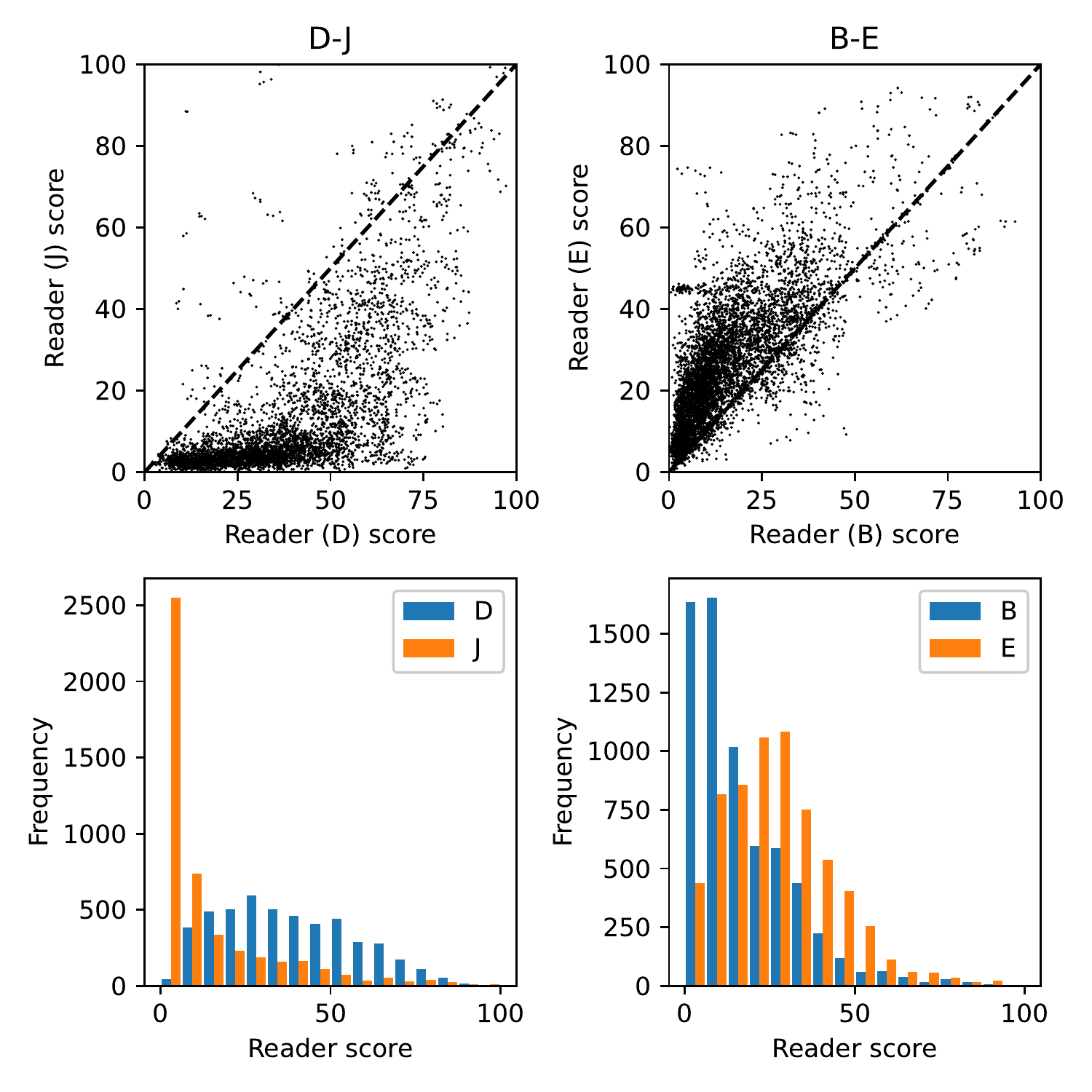}
\end{center}
\caption 
{ \label{fig:figure6}
Pairs of reader scores for two examples: D-J and B-E. The top plots show direct comparisons of the reader scores showing the large variation between the pairs of reader. The bottom plots show histograms of the pairs of reader scores which show the differences in distribution that the readers assess on the images.}
\end{figure}

In Figure~\ref{fig:figure6} we show the variability of individual reader scores for two example pairs. The top row shows the image scores for one reader against another reader, with the reader identifiers in the titles. The bottom row show histograms of the distributions for the pair of readers. The histograms on the bottom row show the significant variation in distribution that the readers show. There is also other substantial variation but separating randomness from more fundamental differences in judgement of density is challenging. 

\begin{figure}[ht] 
\begin{center}
\includegraphics[height=9.8cm]{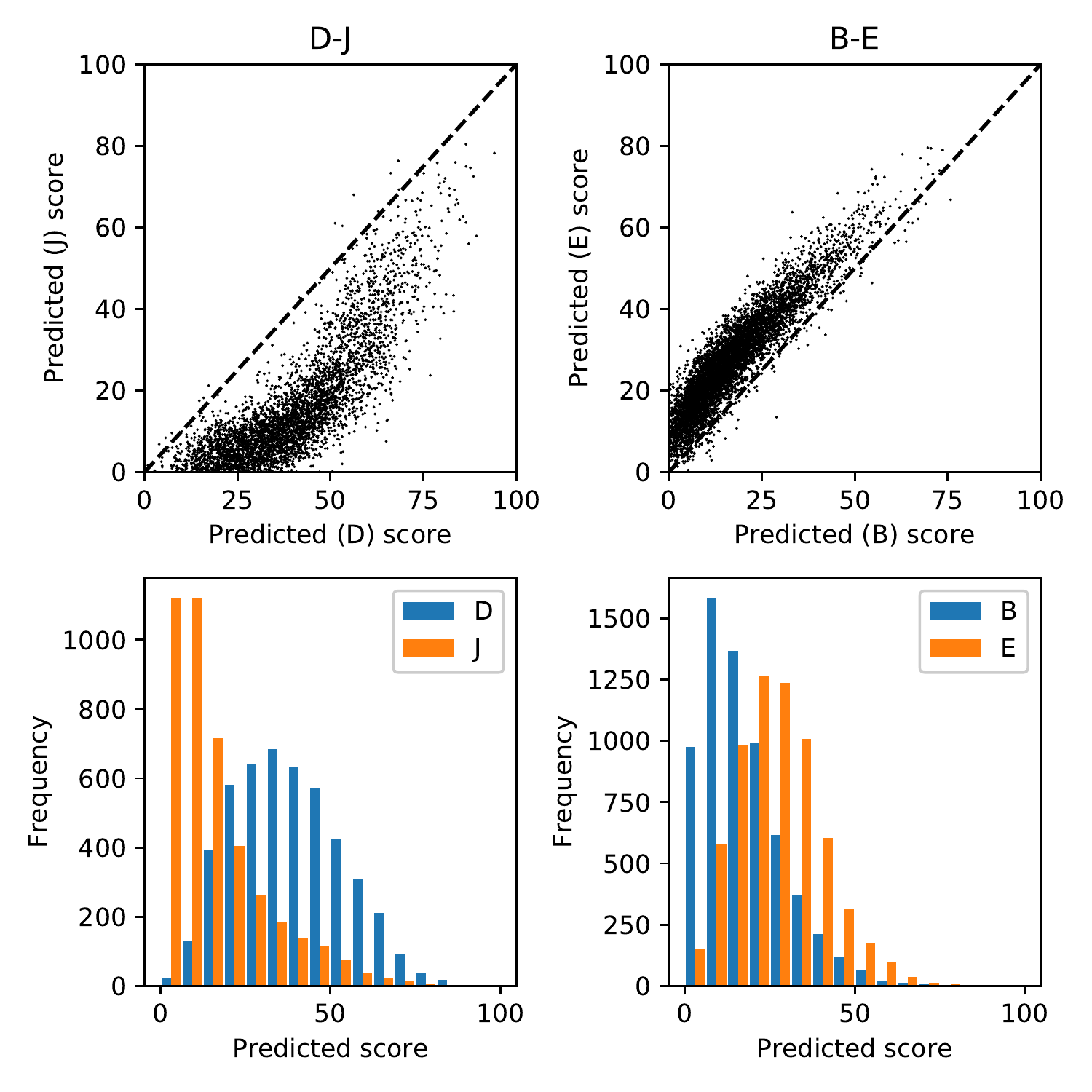}
\end{center}
\caption 
{ \label{fig:figure7}
Pairs of predicted scores for the same pair of examples as in Figure~\ref{fig:figure6}. The models are trained separately on the individual scores of the pair. The top plots show direct comparisons of the predicted scores for the models trained on the different reader scores. The bottom plots show histograms of the pairs of predicted scores which show the differences in distribution due to the differences in labels from the readers.}
\end{figure}

To demonstrate the effect on the model of the label differences shown in Figure~\ref{fig:figure6} we train and validate purely on those images common to the pair but training separate models for each set of reader scores and then plot the predictions in Figure~\ref{fig:figure7}. The top plots are direct comparisons of the predictions found by training and testing on the same images but with the different models trained with the individual labels from the pair of readers. The bottom row shows their prediction distributions. The model predictions and associated reader scores follow the same distribution. The pairs of models also show variation beyond just the distributional differences although substantially less than between the pairs of readers.

\begin{figure}[ht] 
\begin{center}
\includegraphics[height=6.0cm]{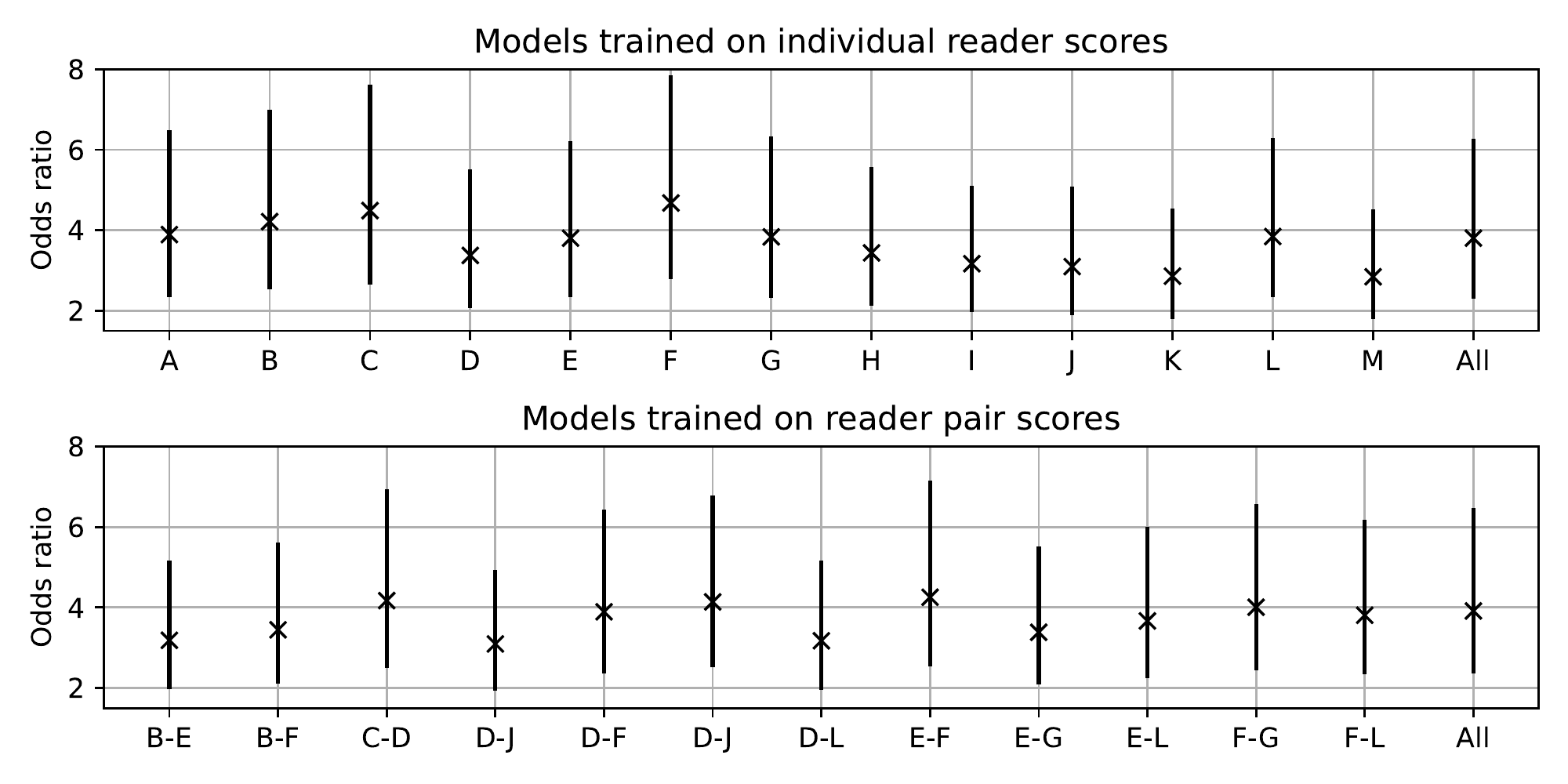}
\end{center}
\caption 
{ \label{fig:figure10}
Odds ratios between the highest density quintile and lowest density quintile for the models trained on individual scores (top) and  paired average (bottom) as well as on the whole data-set. Higher values denote a stronger relationship between predicted density and cancer risk.}
\end{figure}

In previous work the relationship between the averaged VAS scores and cancer risk was shown to be high~\cite{astley2018comparison}. The key metric is the odds ratio between the highest and lowest density quintiles. The higher the value the stronger the relationship between the density predictions and risk of developing breast cancer. In Figure~\ref{fig:figure10} we show plots of these odds ratios when training on the individual and paired reader scores. Every model trained on each individual or paired set of scores show a significant relationship with cancer risk.

\FloatBarrier

\subsection{Method 2: end-to-end deep models to investigate the representation formed}

We now consider the effect of the label variability on the quality of the representation formed by our two end-to-end models. In Table~\ref{tab:table1} we show results for the \textit{single-predictor} and \textit{multi-predictor} (labelled as \textit{Single} and \textit{Multi} respectively). We also show the equivalent results found by using the representation formed by the same ResNet model~\cite{he2016deep} trained on ImageNet~\cite{deng2009imagenet} with no training on the mammography data-set (labelled as \textit{Pre-trained}). Uncertainties were found via bootstrapping with 1000 repeats and taking the 95\% differences. The labels refer to the way we trained and tested the linear mapping: \textit{Av} means we used the entire data-set with labels averaged across the pair of readers; for \textit{Ind 1} and \textit{Ind 2} we trained the linear mapping on individual reader scores and tested on the averaged scores. For \textit{Ind 1} we averaged the metrics across the readers and for \textit{Ind 2} we concatenated all the predictions together and calculated metrics from the total set.  

The improvement by performing fine-tuning from the pre-trained model is substantial and statistically significant for both the \textit{multi-predictor} and \textit{single-predictor} models. The differences between the single and multi-predictors are small. This demonstrates that the representation formed by the pre-trained models on ImageNet can be significantly improved by adaptation to the mammography domain. However, we see little difference between training on averaged labels using the \textit{single-predictor} or the individual labels using the \textit{multi-predictor}. The implication of this is that the representation formed by both end-to-end models is similar, at least in their ability to produce good representations for a linear mapping to the labels.

\begin{table}[ht] 
\centering
\begin{tabular}{|ll|cc|}
\hline
Labels & Model & Rank Corr. & RMSE \\
\hline
Av & Pre-trained & $0.799\pm0.006$ &   $9.88\pm0.12$  \\
Av & Single        & $0.846\pm0.006$ &   $8.63\pm0.12$ \\
Av & Multi         & $0.850\pm0.005$ &   $8.50\pm0.12$ \\
\hline
Ind 1 & Pre-trained &  $0.786\pm0.022$ &   $9.92\pm0.50$ \\
Ind 1 & Single     &  $0.850\pm0.018$ & $8.47\pm0.48$ \\
Ind 1 & Multi      &  $0.854\pm0.018$ &   $8.30\pm0.48$ \\
\hline
Ind 2 & Pre-trained & $0.842\pm0.004$ & $10.33\pm0.12$ \\
Ind 2 & Single     & $0.884\pm0.004$ & $8.96\pm0.13$ \\
Ind 2 & Multi      & $0.888\pm0.003$ & $8.79\pm0.14$ \\
\hline
\end{tabular}
\caption{Metrics for the differences of the model types (pre-trained, single and multi) for the three different label comparisons considered (\textit{Av}., \textit{Ind 1}, \textit{Ind 2}). Both the single-predictor and multi-predictor models show statistically significant improvements from the models trained on ImageNet alone but show minor differences in performance compared to one another}
\label{tab:table1}
\end{table}

\begin{figure}[ht]
\centering
\includegraphics[scale=0.66]{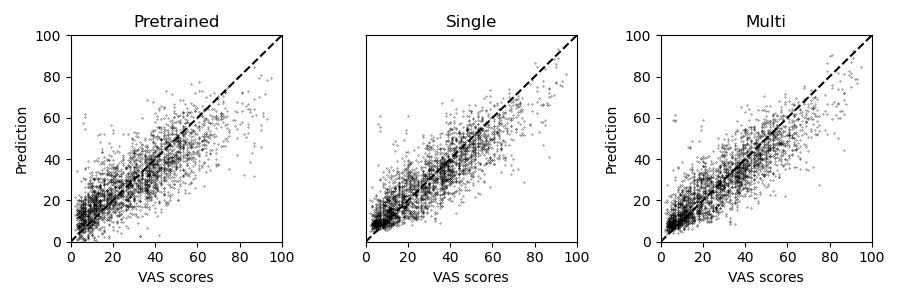}
\caption{Plots of model predictions against reader VAS scores for linear predictions trained on the features produced from the three models.}
\label{fig:dataPlots1a}
\end{figure}

In Figure~\ref{fig:dataPlots1a} we demonstrate the general improvement in performance by finetuning on the pre-trained weights. For one of the readers, we take the three sets of model predictions and show against the reader VAS scores. We see generally a stronger relationship between the reader VAS scores and the predicted scores for the trained models.

To further investigate the similarities between the representations, in Table~\ref{tab:table2} we show the differences in prediction for the single- and multi-predictors. As we have utilised the same mapping function (a linear regression) between the representations in feature space of the two models the differences we are looking for will come only from the representation differences. These results implies the representation formed, with respect to the final density score, are extremely similar once we have trained the models on the mammography images.  

\begin{table}[ht] 
\centering
\begin{tabular}{|ll|cc|}
\hline
Labels & Comparison &  Rank Corr. & RMSE \\
\hline
Av     & M\_S & $0.991\pm0.001$&	$1.98\pm0.04$ \\
Av & M\_Pre & $0.931\pm0.003$	&$5.32\pm0.10$ \\
Av & S\_Pre & $0.933\pm0.003$	&$5.17\pm0.09$ \\
\hline
Ind. 1 & M\_S & $0.984\pm0.002$	& $2.35\pm0.13$ \\
Ind. 1 & M\_Pre & $0.908\pm0.011$ &	$5.66\pm0.31$ \\
Ind. 1 & S\_Pre & $0.906\pm0.011$ &	$5.80\pm0.31$ \\
\hline
Ind. 2 & M\_S & $0.991\pm0.001$ & $2.31\pm0.03$ \\
Ind. 2 & M\_Pre & $0.943\pm0.002$&	$5.59\pm0.07$ \\
Ind. 2 & S\_Pre & $0.941\pm0.002$&	$5.76\pm0.07$ \\
\hline
\end{tabular}
\caption{Metrics for the prediction differences between the pre-trained (Pre), \textit{single-predictor} (S) and \textit{multi-predictor} (M) models.}
\label{tab:table2}
\end{table}

\begin{figure}[ht]
\centering
\includegraphics[scale=0.66]{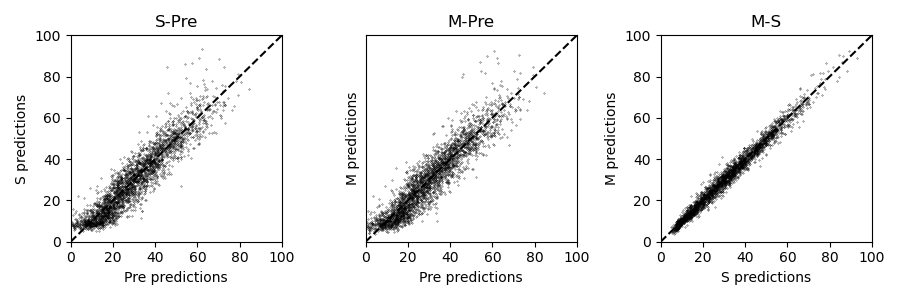}
\caption{Plots of model predictions against other model predictions. Left) \textit{single-predictor} (S) against pretrained (Pre), middle) \textit{multi-predictor} (M) against pretrained, right) \textit{multi-predictor} against \textit{single-predictor}}
\label{fig:dataPlots1b}
\end{figure}

In Figure~\ref{fig:dataPlots1b} we show the similarity in prediction between the three different models for one example set of reader scores. The left plot is the single predictor (S) against the pre-trained (Pre) model, the middle plot is the multi-predictor (M) against the pre-trained model and the right plot is the multi-predictor against the single-predictor. The similarity between the predictions made between the single and multi-predictor models is very high, much more so that between either of the trained models and the pre-trained model.

In Table~\ref{tab:table3} we show the dissimilarity between two runs of the multi and single predictor models compared to their repeated runs - with the repeated runs having identical hyper-parameters sets. These results give empirical examples of the differences in predictions due to different training runs of the same model. The differences between different runs of the single predictor are actually larger than the differences between the single predictor and the multi-predictor (Table~\ref{tab:table2}) which is likely due to the high consistency we are seeing with the multi-predictor. It may be notable that the multi-predictor appears to be more consistent with repeats of the training.

\begin{table}[ht] 
\centering
\begin{tabular}{|ll|cc|}
\hline
Labels & Comparison & Corr. & RMSE  \\
\hline
Av &S\_S    &  $0.989\pm0.001$ & $2.14\pm0.04$ \\
Av &M\_M    &   $0.996\pm0.000$ & $1.30\pm0.03$ \\
\hline
Ind 1& S\_S &  $0.983\pm0.002$ &	$2.51\pm0.15$ \\
Ind 1& M\_M &  $0.991\pm0.001$ & $1.63\pm0.10$ \\
\hline
Ind 2 &S\_S &  $0.989\pm0.001$ & $2.47\pm0.03$ \\
Ind 2 &M\_M &  $0.996\pm0.001$ & $1.59\pm0.02$ \\
\hline
\end{tabular}
\caption{Metrics for the prediction differences between the repeated runs of the multi- (M) and single- (S) predictors individually.}
\label{tab:table3}
\end{table}

\begin{figure}[ht]
\centering
\includegraphics[height=3.5cm]{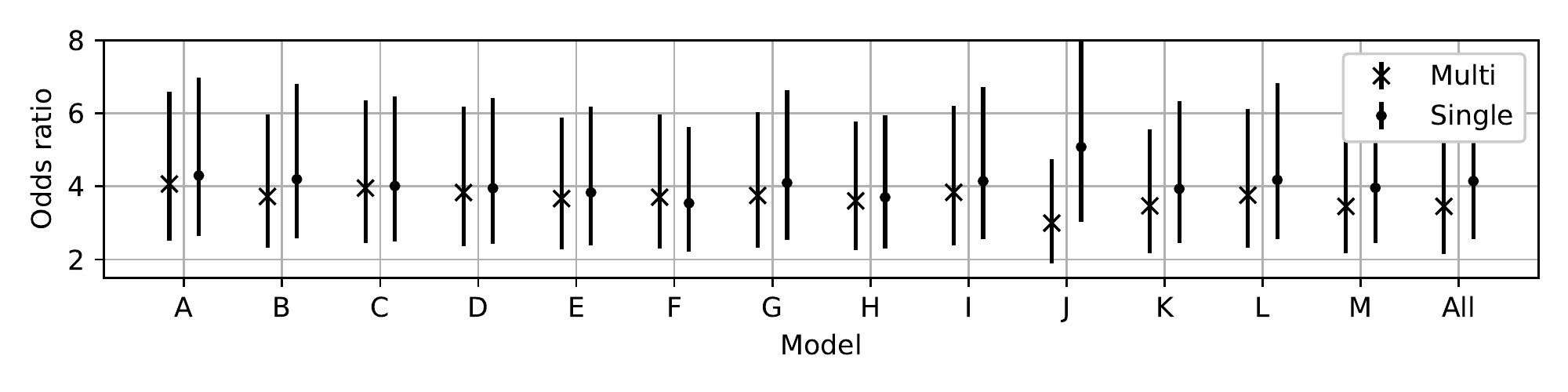}
\caption{Odds ratios for the two models. There are no significant differences between the models in terms of their ability to make risk predictions.}
\label{fig:figure3_model2}
\end{figure}

In Figure~\ref{fig:figure3_model2} we show the cancer odds ratios for the two models with predictions made via the individual models, labelled A-M, and the model trained on all the data (All). Similarly to the results in method 1, we see consistently good ability to make risk predictions.

\section{Discussion}

Breast density is an important risk factor for cancer risk and expert assessed scores have been shown to have a strong relationship with future breast cancer development. Accurate and consistent automated breast density prediction can be employed for masking risk estimation~\cite{mainprize2019prediction} or cancer risk prediction models~\cite{brentnall2015mammographic}.

While the variability of reader labels has been investigated~\cite{sergeant2013same,sperrin2013correcting}, in this paper we showed the effect of this variability on deep learning based predictive models. Deep learning models can be considered to consist of two parts, one which forms the representation and a second part that performs the mapping to the final prediction. We investigated the effect of the label variability on the representation formed by the model and, separately, on the mapping from the representation to the prediction. We also considered the effect of reader density score differences on the quality of the models and on the ability to judge the quality of the models. 

We showed that label variability has a considerable impact in terms of the assessment of the quality of the models. When we compared the predictions of the feature extraction linear regression models (Method 1) to averaged density scores we find better metrics than comparing to individual density scores. We also showed that when we train the linear mapping on labels with the variability in the labels reduced, that our results improve considerably. 

The mapping of the vector representations to the final prediction is affected considerably by the variability in reader scores. All mappings show good risk prediction scores independently but combining predictions together from models trained on different datasets would need to be done with considerable care due to the different distributions of predictions produced by the models.

To investigate the effect on the representation we made a novel alteration to a standard deep network which enables us to train on the individual reader scores. The representation formed by this approach was compared to that of a deep network trained on the averaged labels by applying linear regression to map the representations to the labels. 

Both of the models trained end-to-end on the mammograms do significantly better than the mappings applied to the representations formed by training on models only on ImageNet. The representations formed by the two different models produce similar final mapped predictions. In fact the differences between the predictions are comparable to the differences that occur due to performing multiple training runs of the same model. 

As our fine-tuned models both perform significantly better than ImageNet trained models, and produce very similar outputs to each other, this implies that the models must both be finding signal in the images which are not being significantly affected by label variability. This may have important implications for the deployment of these types of models. 

Models trained to make predictions using different labelled data-sets may not be usable together due to the significant differences we see in the mapping of our representations to the final data. However, if the representations that are formed are consistent then it may be possible to transfer different trained models formed to produce representations. Mappings from representations do not require as much data to be effectively trained as the formation of the representation. It may be possible to train representation forming models on large data-sets and then deploy them to have their final mappings assessed on much smaller local data-sets. This may enable models trained on one type of mammography data-sets, one example would be images from a certain manufacturer or machine type, being utilisable on different types of data-sets. 

\section{Conclusions}

In this paper we have shown that variability in reader breast density labels has a sizeable effect on the final mapping from representation space to density prediction. The consequences of this are that models trained on different data-sets cannot be meaningfully used together if the direct comparison between individuals assessed by the different models are required. However, the risk scores, when considered on their own without comparison to other model predictions, are all good.

The quality of the representation formed when deep learning models are trained on the variable data appear to be similar and not significantly affected by the variability in the labels. If the variability in the labels had a significant effect on the model representation we would expect to see significant differences in the outputs of the two models when we applied linear regression to them. However, we see very similar results, implying that the representations formed by both models are, in some sense, extremely similar. We may see less variability in repeated training runs of the multi-predictor approach which may imply it is more stable in training. The similarity in quality of the representation of the two trained deep models provides evidence that there is a strong density-related signal in the images which the models are easily able to learn. 

We conclude that the deep learning representations formed by training on these types of labels are similar regardless of the method used. However, the final mapping from representation space has to be considered very carefully to ensure that false conclusions are not being drawn.

\section*{Disclosures}
The authors declare no conflicts of interest. Ethics approval for the PROCAS study was through the North Manchester Research Ethics Committee (09/H1008/81). Informed consent was obtained from all participants on entry to the PROCAS study.

\section*{Acknowledgements}
D. Gareth Evans, Elaine Harkness and Susan M Astley are supported by the National Institute for Health Research (NIHR) Manchester Biomedical Research Centre (IS-BRC-1215-20007).

\bibliographystyle{unsrt}
\bibliography{bibfile}
\FloatBarrier

\end{document}